\documentclass[10pt,twocolumn,letterpaper]{article}

\usepackage{iccv}
\usepackage{times}
\usepackage{epsfig}
\usepackage{graphicx}
\usepackage{amsmath}
\usepackage{amssymb}
\usepackage{float}
\usepackage{multirow}
\usepackage{booktabs}

\usepackage[breaklinks=true,bookmarks=false]{hyperref}

\iccvfinalcopy 

\def\iccvPaperID{1983} 

\ificcvfinal\fi

\begin{document}

\title{Generative Modeling for Small-Data Object Detection}

\author{
    Lanlan Liu$^{1,2}$\thanks{This work was conducted when Lanlan Liu was an intern at Google.} \quad Michael Muelly$^{2}$ \quad Jia Deng$^{3}$ 
 \quad Tomas Pfister$^{2}$\thanks{Corresponding author.} \quad Jia Li$^{4}$\\
 $^{1}$University of Michigan, Ann Arbor 
 \quad $^{2}$Google Cloud AI \quad $^{3}$Princeton University \quad $^{4}$Stanford University\\\
 {\tt\footnotesize{llanlan@umich.edu}  \tt\footnotesize{mmuelly@google.com}  \tt\footnotesize{jiadeng@cs.princeton.edu} \tt\footnotesize tpfister@google.com} \tt\footnotesize{lijiali@cs.stanford.edu}
}

\maketitle
\ificcvfinal\thispagestyle{empty}\fi

\begin{abstract}
This paper explores object detection in the small data regime, where only a limited number of annotated bounding boxes are available due to data rarity and annotation expense.
This is a common challenge today with machine learning being applied to many new tasks where obtaining training data is more challenging, \eg in medical images with rare diseases that doctors sometimes only see once in their life-time.
In this work we explore this problem from a generative modeling perspective by learning to generate new images with associated bounding boxes, and using these for training an object detector.
We show that simply training previously proposed generative models does not yield satisfactory performance due to them optimizing for image realism rather than object detection accuracy.
To this end we develop a new model with a novel unrolling mechanism that jointly optimizes the generative model and a detector such that the generated images improve the performance of the detector.
We show this method outperforms the state of the art on two challenging datasets, disease detection and small data pedestrian detection, improving the average precision on NIH Chest X-ray by a relative 20\% and localization accuracy by a relative 50\%.
\end{abstract}

\section{Introduction}

Generative Adversarial Networks (GANs)~\cite{goodfellow2014generative} have recently advanced significantly, with the latest models~\cite{anonymous2019large, karras2018style} being able to generate high quality photo-realistic images that are almost indistinguishable from real images. 
A natural question that has recently started being explored~\cite{li2017perceptual,ouyang2018pedestrian,shrivastava2017learning} is whether these generated images are useful in some other ways; for example, could they be useful training data for downstream tasks?

One common computer vision task that could benefit from generated data is object detection~\cite{lin2018focal,renNIPS15fasterrcnn} which currently requires a large amount of training data to obtain good performance.
But for many object detection tasks, large datasets are difficult to obtain due to rare objects and difficulties in obtaining object location annotations. 
One common example is with medical images -- disease detection has very little labeled object bounding box data because the diseases by nature are rare, and annotations can only be done by professionals, and thus are costly. 
Solving such rare data object detection problems is valuable: for example, for disease localization, a good disease detector can help provide assistance to radiologists to accelerate the analysis process and reduce the chance of missing tumors, or even provide a medical report directly if a radiologist is not available.

In this paper we explore using generative models to improve the performance in small-data object detection. 
Directly applying existing generative models is problematic. 
First, previous work on object insertion for generative models often needs segmentation masks, which are often not available \eg in disease detection tasks. 
Second, GANs are designed to produce realistic images (indistinguishable from real images), but realism does not guarantee that it can help with the downstream object detection task.
In particular, there is no direct feedback from the detector to the generator; which means the generator cannot be trained explicitly to improve the detector. 

\begin{figure}[!t]
\begin{center}
\includegraphics[width=1\linewidth]{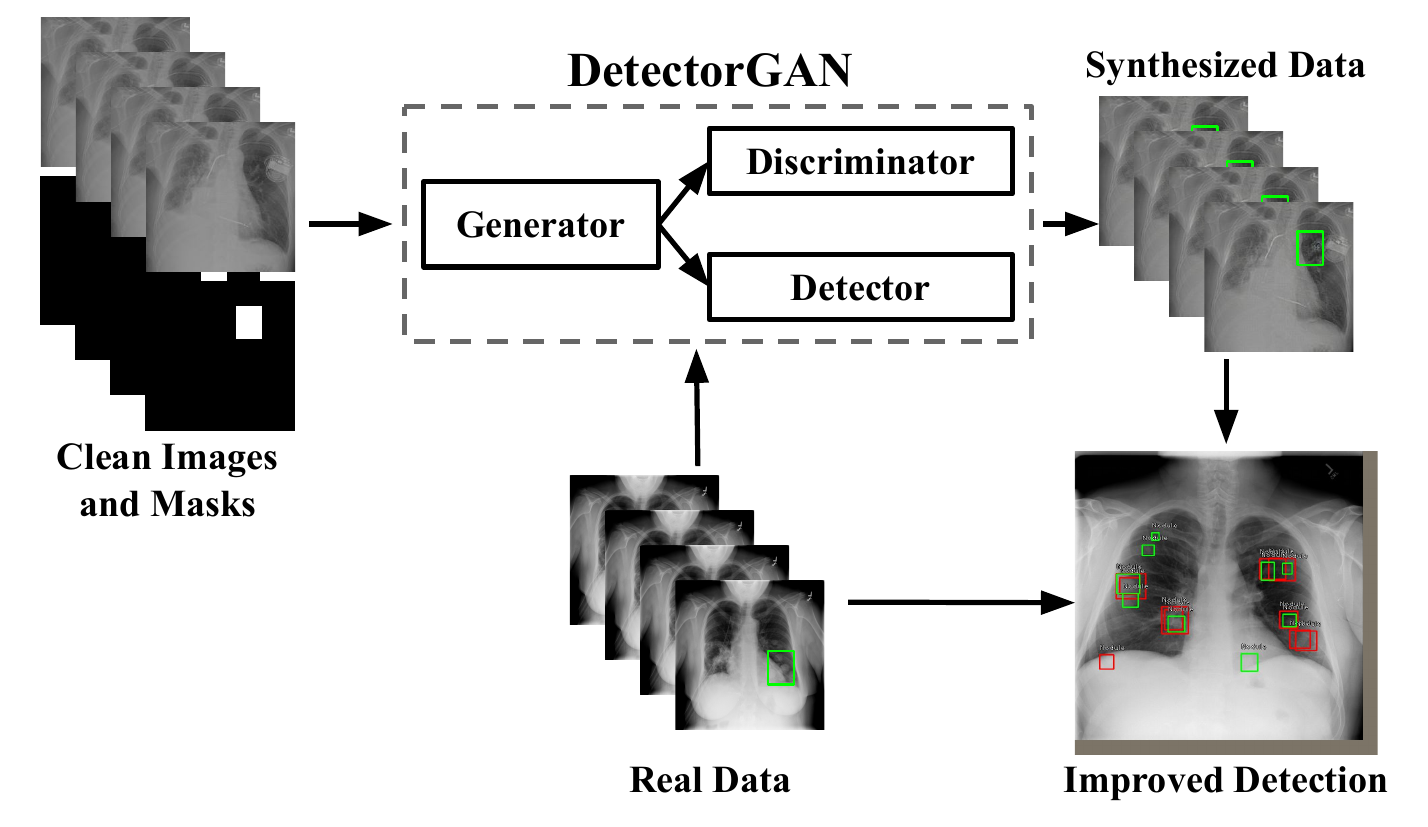}
\end{center}
   \caption{ 
   DetectorGAN generates object-inserted images as synthesized data to improve the detection performance. DetectorGAN integrates a detector into the generator-discriminator loop. 
   }
\label{fig:overview}
\end{figure}

To address this, we propose a new DetectorGAN model (shown in Fig.~\ref{fig:overview}) that connects the detector and the GAN together. 
This joint model integrates a detector into the generator-discriminator pipeline and trains the generator to explicitly improve the detection performance.

DetectorGAN has two branches after the generator: one with discriminators to improve realism and interpretability of the generated images, and another with a detector to give feedback on how well the generated images improve the detector.
We jointly optimize the adversarial losses and detection losses. 
To generate images that are beneficial for the detector, the loss formulation is non-trivial. 
One difficulty is that our goal is for the generated images to improve the detector performance of real images, but the generator cannot receive gradients from the detection loss on real images because the real images are not generated. 
To address this, the proposed method bridges this link between the generator and the detection loss on real images by unrolling one  forward-backward pass of the detector training.

We demonstrate the effectiveness of using DetectorGAN to improve small-data object detection in two datasets for disease detection and pedestrian detection. 
The detector-integrated GAN model achieves state of the art performance on the NIH chest X-ray disease localization task, benefiting from the additional generated training data.
In particular, DetectorGAN improves the Average Precision of the nodule detector by a relative 20\% by adding 1000 synthetic images, and outperform the state of the art on localization accuracy by a relative 50\%.
We also show that the proposed framework significantly improves the quality of the generated images: a radiologist prefers generated images by DetectorGAN over alternative methods in 96\% of cases.
The detector model can be integrated into almost any existing GAN models to force them to generate images that are both realistic and useful for downstream tasks. 
We give the pedestrian detection task and the associated PS-GAN~\cite{ouyang2018pedestrian} as an example, demonstrating a significant quantitative and qualitative improvement in the generated images.

Our contributions are:
\begin{enumerate}
\item To the best of our knowledge, this work is first to integrate a detector into the GAN pipeline so that the detector gives direct feedback to the generator to help generate images that are beneficial for detection.
\item We propose a novel unrolling method to bridge the gap between the generator and the detection performance on real images.
\item The proposed model outperforms GAN baselines on two challenging tasks including disease detection and pedestrian detection, and achieves the state-of-the-art performance on NIH chest X-ray disease localization.
\item We are the first few works to explore GANs with downstream vision tasks such as small-data object detection.
\end{enumerate}

\section{Related Work}

\paragraph{Image-to-image Translation.}
Based on a conditional version of Generative Adversarial Networks (GANs)~\cite{goodfellow2014generative}, Isola et al~\cite{isola2017image} pioneered the general image-to-image translation task.
Afterwards multiple other works have also exploited pixel-level reconstruction constraints to transfer between source and target domains~\cite{zhang2017stackgan, wang2018pix2pixHD}.
These image-to-image translation frameworks are powerful, but require training data with paired source/target images, which are often difficult to obtain.
Unpaired image-to-image translation frameworks~\cite{zhu2017unpaired,liu2017unsupervised,shrivastava2017learning,DBLP:conf/icml/KimCKLK17,lee2018unsupervised} remove this requirement of paired-image supervision; in CycleGAN~\cite{zhu2017unpaired} this is achieved by enforcing a bi-directional prediction between source and target. The proposed DetectorGAN falls in the category of unpaired image-to-image translation frameworks. Its novelty is that it integrates a detector into GAN to generate images as training data for object detection.

\paragraph{Object Insertion with GANs.} 
The idea of manipulating images by GANs has been explored recently~\cite{lee2018context,hong2018learning,chien2017detecting,lin2018st,ouyang2018pedestrian, lee2019inserting, lin2018st}. 
These works use generative models to edit objects in the scene. 
In contrast, (1) our method doesn't require any segmentation information; and (2) our goal is to gain quantitative improvement on object detection task while prior works focus on qualitative improvement such as realism.

\paragraph{Integration of GANs and Classifiers.}
Beyond the basic idea of using adversarial losses to generate realistic images, some GAN models integrate auxiliary classifiers into the generative model pipeline, such as Auxiliary Classifier GAN (ACGAN) and related works \cite{odena2017conditional,bazrafkan2018versatile,dash2017tac,bousmalis2017unsupervised, hoffman2018cycada}. 
At a first glance, these models bear some similarity with our integration with detector. 
However, we differ from them both conceptually and technically. 
Conceptually, these methods only improve the realism of the generated images and have no intention to improve the integrated classifier; in contrast, the purpose of our integration is to improve the detection performance. 
Technically, our loss formulation is different: ACGAN minimizes classification losses only on synthetic images and has no guarantee for improving performance on real images, whereas ours optimizes losses on both synthetic and real by adding unrolling step.
Nevertheless, we construct a baseline with ACGAN-like losses, which only minimizing detection losses on synthetic images, and show that our proposed method outperforms it.

\paragraph{Data Augmentation for Object Detection} 
There are some works using data augmentation to improve object detection.
A-Fast-RCNN~\cite{WangCVPR17afrcnn} uses adversarial learning to generate hard data augmentation transformations, specifically for occlusions and deformations. 
It differs from the method in this paper in two major ways: (1) It is not a GAN model -- it does not generate images but instead adds adversarial data augmentation into the detector network. 
In contrast, our model has a discriminator and detector that work together to generate synthetic images. 
(2) Its goal is to `learn an object detector that is invariant to occlusions and deformations'.
In contrast, this paper focuses on generating synthetic data for the problem setting where the amount of training data is limited.

Perceptual GAN~\cite{li2017perceptual} generates synthetic images to improve the detection. 
However, it is designed specifically for small-sized object detection by super-resolving the small-sized objects into better representations.
Their method does not generalize to general object detection. 

Concurrent unpublished work PS-GAN~\cite{ouyang2018pedestrian} is most closely related: synthetic images are generated to improve pedestrian detection.  
They generate synthetic images using a traditional generator-discriminator architecture. 
In contrast, we add a detector in the generator-discriminator loop and have direct feedback from the detector to the generator.

\section{DetectorGAN}

Our DetectorGAN method generates synthetic training images to directly improve the detection performance.
It has three components: a generator, (multiple) discriminators, and a detector. 
The detector gives feedback to the generator about whether the generated images are improving the detection performance. 
The discriminators improve the realism and interpretablity of the generated images; that is, the discriminators help to produce realistic and understandable synthetic images. 

\subsection{Model Architecture}
We implement our architecture based on CycleGAN~\cite{zhu2017unpaired}. 
The generator in DetectorGAN generates synthetic labelled (object-inserted) images that are fed into two branches later: the discriminator branch and the detector branch. 
We consider clean images without objects belong to domain X, and labelled images with objects belong to domain Y.

\paragraph{Generators.}
We use a ResNet generator with 9 blocks as our generators $G_X$ and $G_Y$ following \cite{zhu2017unpaired,he2016deep}. The forward generator $G_X$ takes two inputs: one is a real clean image, which is used as the background image to insert objects. 
The other one is a mask where the pixels inside the bounding box of the object to insert are filled with ones while the rest are zeros. 
The output of the generator is a synthetic image with the input background and an object inserted at the marked location.
Inversely, the backward generator $G_Y$ takes a real labelled image and a mask showing the object location, and outputs an image with the indicated object removed.

Plausible inserting locations of objects are difficult to obtain. 
In this paper, for the NIH disease task, we obtain these locations by pre-processing and random sampling. 
In theory, the location could be in any position in the lung area, but since in practice we do not have segmentation mask for the lung area, we first match each clean image to the most similar labelled image with bounding box and then randomly shift the location around to get the sampled ground-truth box location. 
For the pedestrian detection task, we follow the setup in the previous work~\cite{ouyang2018pedestrian}. 
It is notable that the selection of mask locations does not change our method -- as an alternative one could use trainable methods to predict plausible locations.

\paragraph{Discriminators.}
Our method contains two global discriminators $DIS_{globalX}$ and $DIS_{globalY}$ as in Cycle-GAN\cite{zhu2017unpaired}, and a local discriminator $DIS_{localX}$ for local area realism~\cite{lee2019inserting,li2017generative}. 
The global discriminator $DIS_{globalX}$ and the local discriminator $DIS_{localX}$ discriminates between real labelled images and synthetic labelled images (generated by $G_X$), globally on the whole images or locally on the bounding box crops. 
$DIS_{globalY}$ discriminates what $G_Y$ generates (synthetic clean images by removing objects from real labelled images) and real clean images. 
$DIS_{localY}$ is not needed because conceptually we do not care much about the local realism after removing an object.
We use $70\times70$ PatchGAN following \cite{johnson2016perceptual,isola2017image,zhu2017unpaired} for all of our discriminators.

\paragraph{Detector.}
The detector $DET$ takes both real and synthetic labelled images with objects as input and outputs bounding boxes. 
In our implementation we use the RetinaNet detector~\cite{lin2018focal}. 
But we are not only limited to RetinaNet: as long as the detector is trainable, we can integrate it into the loop.

\subsection{Train Generator with Detection Losses}

The objective of the generator $G_X$ is to generate images with objects inserted that are both realistic and beneficial to improve object detection performance. 
One of our main contributions is that we propose a way to backpropagate the gradients derived from detection losses back to the generator to help the generator to generate images that can better help improve the detector. 
In other words, the detection losses give the generator feedback to generate useful images for the detector.

We note the detection loss (regression and classification losses) as $L_d(\cdot)$, where $\cdot$ is a labelled image, either real or synthetic. The detection loss on real images and synthetic images are:
\begin{equation}
\begin{split}
L_{Det}^{real}(DET) = E_{y \sim p_{data}(Y)}[L_{d}(DET(y))] \\
\label{eqn:loss_detb}
\end{split}
\end{equation}

\begin{equation}
\begin{split}
L_{Det}^{syn}(G_X, DET) = E_{x \sim p_{data}(X)}[L_{d}(DET(G_X(x)))] \\
\label{eqn:loss_deta}
\end{split}
\end{equation}

\paragraph{Unroll to Optimize Detection Loss on Real Images.}

\begin{figure}[!t]
\begin{center}
\includegraphics[width=1\linewidth]{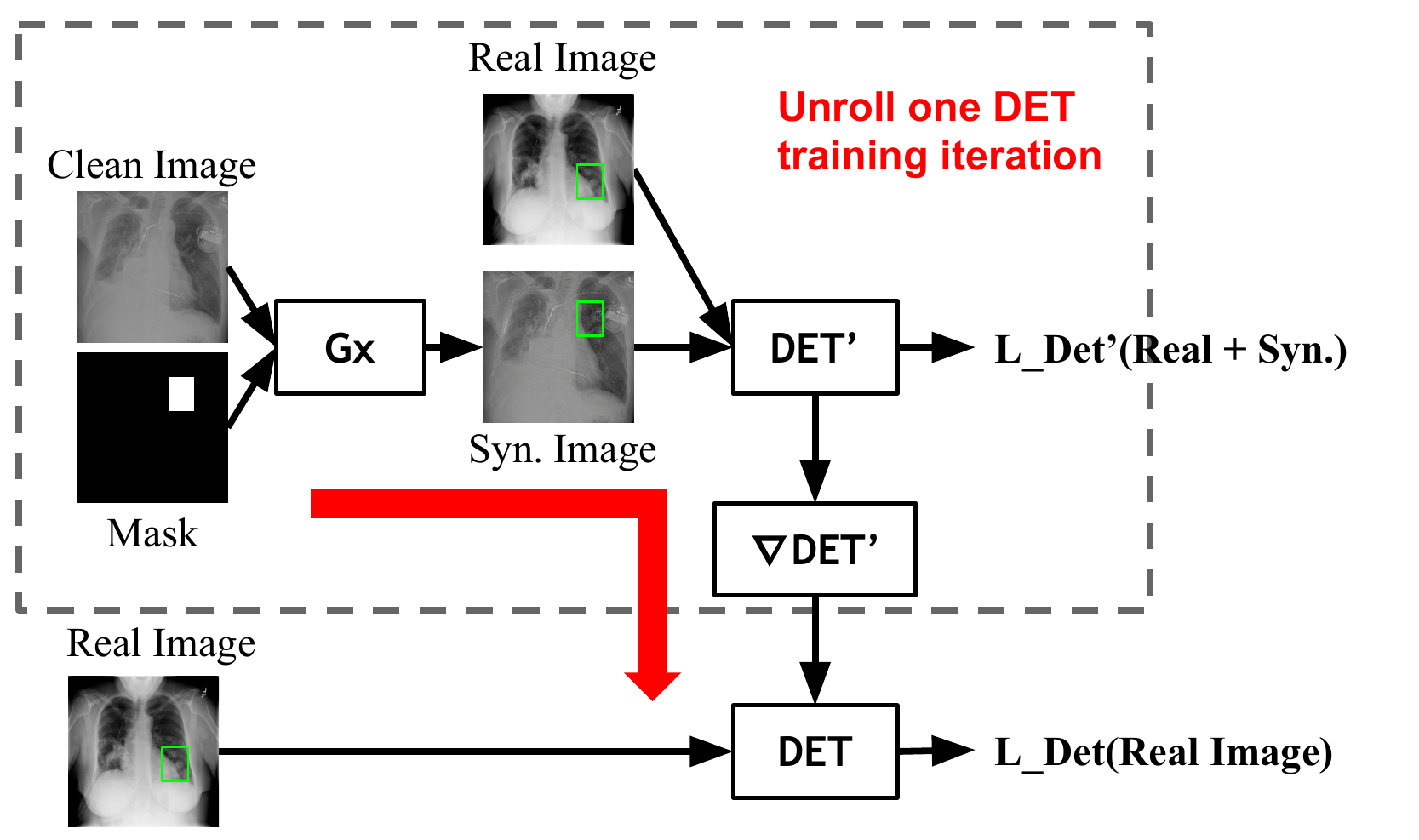}
\end{center}
   \caption{
    The illustration for Eqn.~\ref{eqn:loss_unroll} -- unrolling one forward-backward pass for training $DET$ to bridge the link between $G_X$ and $L_{Det}^{real}$ (detection loss on real images). Detection loss on real images has no direct link to the generator $G_X$. Last step of training old $DET$ (noted as $DET'$ in the figure, refers to same $DET$ module but in the previous training step) is unrolled as in the dotted rectangle. The red arrow represents the fact that there is a differentiable link between $G_X$ and $L_{Det}^{real}$ after the unrolling.
   }
\label{fig:method}
\end{figure}

Intuitively, given a real image $y$, the goal of $G_X$ is to use generated images to help minimize the detection loss on real images. 
That is, $G_X$ should be trained to minimize the loss $L_{Det}^{real}$ in Eqn.~\ref{eqn:loss_detb}.
However, there is no $G_X$ involved at the first glance -- the loss $L_{Det}^{real}$ does not depend on the weights of the $G_X$ so $G_X$ cannot be trained. 
But we observe that even though there is no direct link in one forward-backward loop from $G_X$ to real images, the detector is trained by synthetic images generated by $G_X$ in the previous step. 
We propose to bridge the link between  $G_X$ and the real image detection loss $L_{Det}^{real}$ by unrolling a single forward-backward pass of the detector as shown in Eqn.~\ref{eqn:loss_unroll}. 
A visualization of this unrolling process is shown in Fig.~\ref{fig:method}. 
This allows us to train $G_X$ with respect to the loss $L_{Det}^{real}$.

\begin{equation}
\begin{split}
\tilde{L}_{Det}^{real}(G_X, DET) = E_{y \sim p_{data}(Y)}[L_{d}(DET(y))] \\
\text{where weights of $DET$, $W_{DET}$, is updated with} \\
\frac{\partial (L_{Det}^{real}(DET) + L_{Det}^{syn}(G_X, DET)))} {\partial W_{DET}}  \\
\label{eqn:loss_unroll}
\end{split}
\end{equation}

Specifically, we train the weights $DET$ with synthetic images and real images for one iteration and obtain the gradients on $DET$.
These gradients are linked to the generated synthetic images and thus to the weights in the generator $G_X$. 
Then we use the updated $DET$ to get the $L_{Det}^{real}$ loss and gradients. 
In this way, we obtain a link from $G_X$ to $DET$ and then to $L_{Det}^{real}$.

Intuitively, this Eqn.~\ref{eqn:loss_unroll} can be seen as a simple estimation of how the change in $G_X$ will change detection performance on real images in Eqn.~\ref{eqn:loss_detb}.

\paragraph{Detection Loss on Synthetic Images.}
The generator aims to make the synthetic images helpful for the detector.
It maximizes the detection loss on synthetic images (Eqn.~\ref{eqn:loss_deta}) 
to generate images that the detector has not seen before and cannot predict well. 
In this case the generated images can help improve the performance. 

One might think the generator should instead minimize the detection loss on synthetic images.
This shares some similar ideas with ACGAN-like losses, where the auxiliary classification loss on synthetic images is minimized to improve realism. 
But for our goal to improve the detection performance on real images, minimizing detection losses on synthetic images may not help, or may even hurt the detection performance on real images. 
The intuition behind this is that synthetic objects may distract away from the optimization goal of the detector. 
In our experiments, we show that minimizing synthetic image losses like ACGAN harms detection performance on real images.

\begin{table*}[!t]
\caption{Nodule AP on expanded annotation setting on the test set with IoU = 0.1 for NIH. Baseline is using only real training data. 
We add 1000 synthetic images from CycleGAN and GAN-D for training. }
\label{tlb:ap}
\begin{center}
\begin{tabular}{lcc}
\hline
Training data & Nodule AP & Nodule Recall\\
\hline\hline
Real data only  &  0.124 & 0.184 \\
Real + syn from ACGAN-like losses & 0.154 & 0.607 \\
Real + syn from CycleGAN + BboxLoss & 0.196 & 0.541  \\
Real + syn from DetectorGAN - unrolling & 0.203 & 0.544 \\
Real + syn from DetectorGAN & \textbf{0.236} & \textbf{0.649} \\
\hline
\end{tabular}
\end{center}
\end{table*}

\begin{table*}[!t]
\caption{Localization accuracy with different $T_{IOU}$ on ``old annotations'' test set for NIH.}
\label{tlb:acc_u}
\begin{center}
\begin{tabular}{lccccccc||c}
\hline
$T_{IOU}$ & 0.1 & 0.2 & 0.3 & 0.4 & 0.5 & 0.6 & 0.7 & Avg \\
\hline\hline
Wang et al. \cite{wang2017chestx} &  0.14 & 0.05 & 0.04 & 0.01 & 0.01 & 0.01 & 0.00 & 0.04 \\
Zhe et al. \cite{li2017thoracic} &  \textbf{0.40}  & 0.29 & 0.17 & 0.11 & 0.07 & 0.03 & 0.01 & 0.15 \\
RetinaNet: real & 0.15 & 0.15  & 0.15 & 0.08 & 0.08 & 0.00 & 0.00 & 0.09 \\
RetinaNet: real + syn from CycleGAN + BboxLoss  & 0.31 & 0.31  & 0.23 & 0.23 & 0.00 & 0.00 & 0.00 & 0.15 \\
RetinaNet: real + syn from DetectorGAN & 0.31 & \textbf{0.31}  & \textbf{0.31} & \textbf{0.23} & \textbf{0.23} & \textbf{0.15} & \textbf{0.08} & \textbf{0.23} \\\hline
\end{tabular}
\end{center}
\end{table*}

\subsection{Overall Losses and Training}

\paragraph{Overall Losses.}
The objective of the generator $G_X$ is to generate images with an object inserted at the indicated location in background images. 
The generated images should be both realistic and beneficial to improve object detection performance. 
In other words, the DetectorGAN model should generate images that: 
(1)~can help to train a better detector;
(2)~have an object inserted; and 
(3)~are indistinguishable from real images, globally and locally.

We have introduced losses to help the detector above. 
For inserting an object, we use an L1 loss to minimize the loss between the synthetic object crop and the real object crop (refered as BboxLoss):

To generate realistic images, we have adversarial losses for the global discriminators and the local discriminator. 
For $DIS_{globalX}$, the adversarial loss is $L_{GAN}(G_X, DIS_{globalX})$ as in Eqn.~\ref{eqn:loss_globalx}. $L_{GAN}(G_Y, DIS_{globalY})$ and $L_{GAN}(G_X, DIS_{localX})$ are similar.

\begin{equation}
\begin{split}
L_{GAN}(G_X, DIS_{globalX}) = E_{y \sim p_{data}(Y)}[\text{log} DIS_{globalX}(y)] \\
 + E_{x \sim p_{data}(X)}[\text{log} (1 - DIS_{globalX}(G_X(x)))]
\label{eqn:loss_globalx}
\end{split}
\end{equation}

In addition, we use cycle consistency losses and identity losses 
to help preserve information from the whole image.

Here $G_X$ and $G_Y$ aim to fool the discriminators while the discriminators aim to discriminate between fake and real images. The generator and discriminators thus optimize
$\min_{G_X, G_Y} \max_{DIS_{globalX}, DIS_{localX}, DIS_{globalY}}$
$[L_{GAN} (G_X, DIS_{globalX}) + L_{GAN} (G_Y, DIS_{globalY}) + L_{GAN} (G_X, DIS_{localX})]$.

We update the weights of the detector $DET$ by minimizing the detection losses for both real images and synthetic images: it minimizes Eqn.~\ref{eqn:loss_detb} and Eqn.~\ref{eqn:loss_deta}.

\paragraph{Training.}
In summary, when updating the discriminators, the goal is to maximize the discriminator losses on generated images and minimize the losses on real nodule images. 
When updating the detector, the goal is to minimize the detection losses for both real and generated nodule images. 
When updating the generator, the goal is to: 
(1)~minimize the discriminator losses on generated images; 
(2)~minimize detection loss on real object images, 
(3)~maximize detection loss on generated images. 

We use a history of synthetic images~\cite{shrivastava2017learning}, and for faster convergence we pretrain the discriminator-generator pair and the detector separately and then train them jointly. 
When we have a labelled image without bounding box annotations, we still update the discriminator $DIS_{globalX}$ to improve global realism.

\begin{figure*}[htbp]
\begin{center}
\includegraphics[width=1\linewidth]{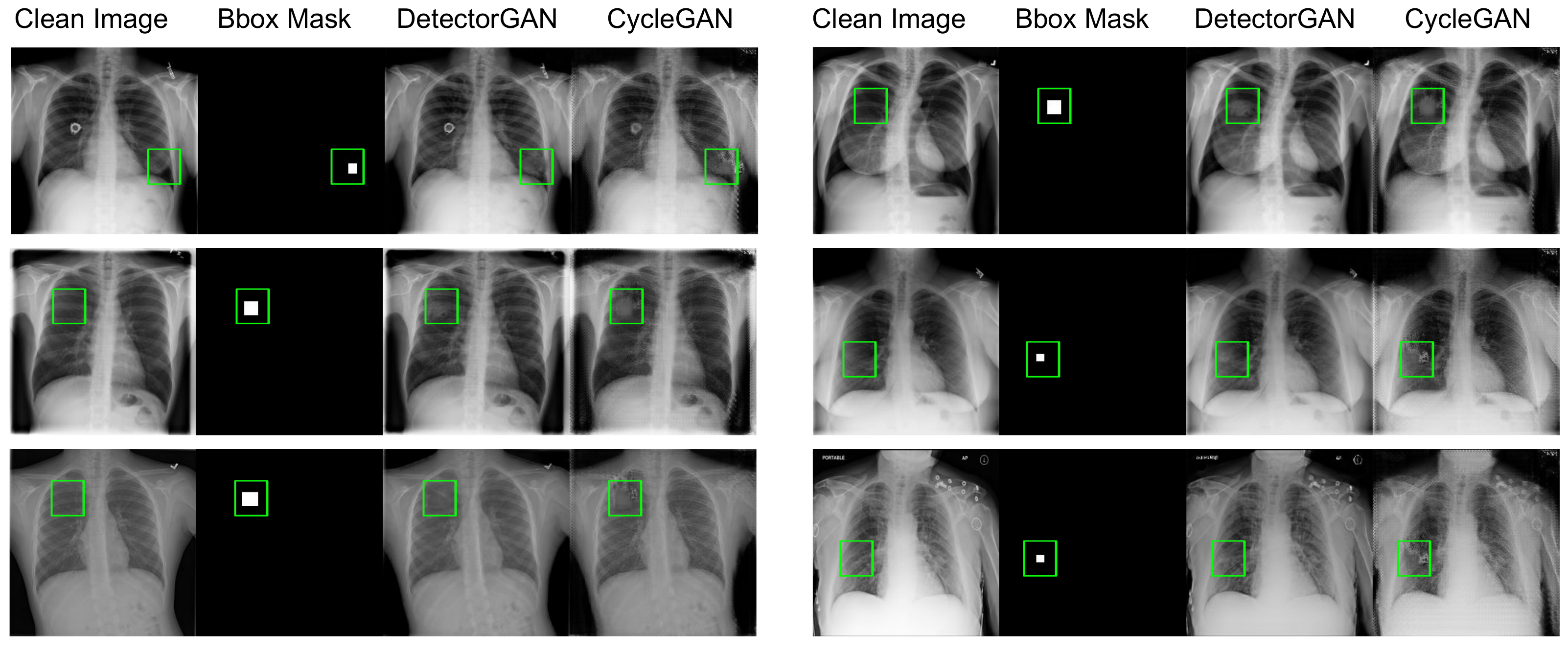}
\end{center}
   \caption{Example generated images from CycleGAN and DetectorGAN for NIH. The details are high-lighted in green boxes (added for visualization).
   Both methods generate synthetic images from clean images and bounding box masks. 
   DetectorGAN generates nodule inserted images with better local and global quality.}
\label{fig:img_quality}
\end{figure*}

\begin{figure*}[htbp]
\begin{center}
\includegraphics[width=1\linewidth]{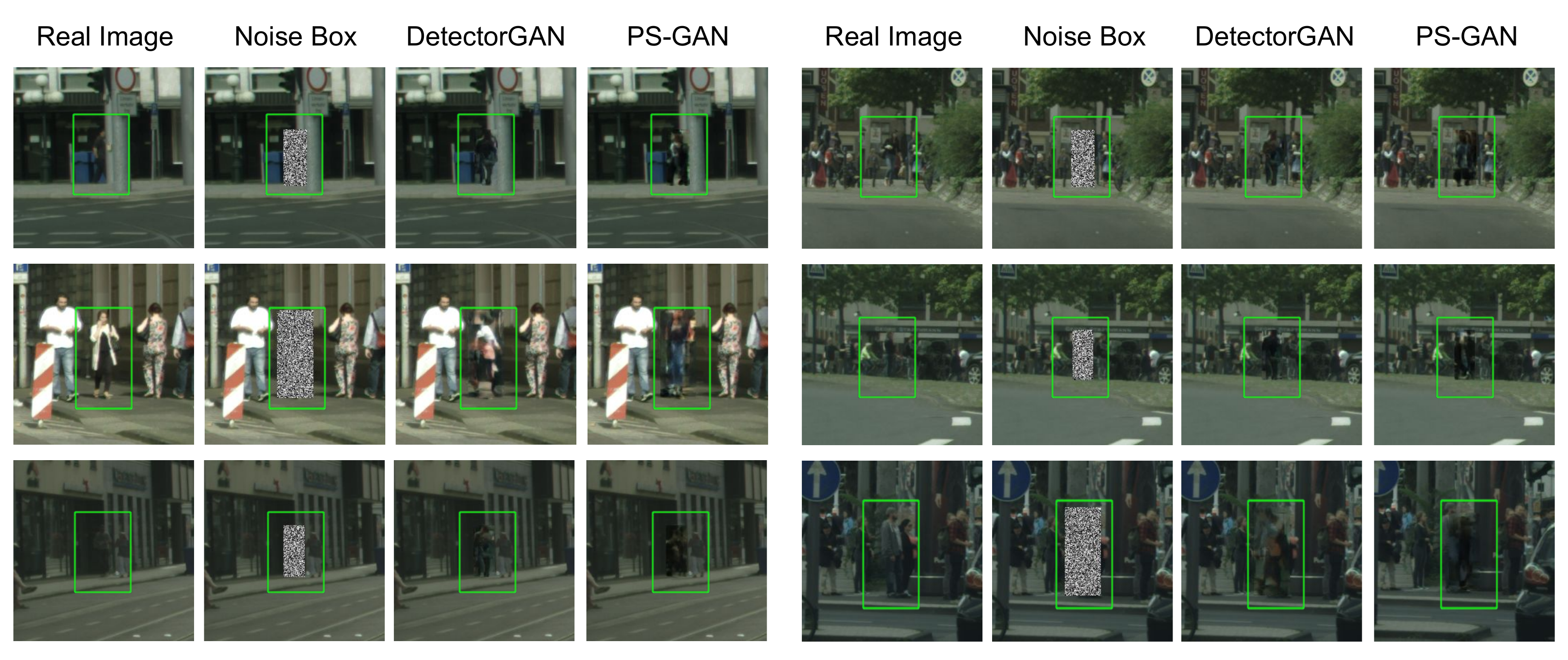}
\end{center}
   \caption{Examples of generated synthetic images from PS-GAN and DetectorGAN. The details are high-lighted in green boxes (added for visualization). We see qualitative improvement for pedestrian task as well.}
\label{fig:ped_img}
\end{figure*}

\begin{table*}[htbp]
\caption{Localization accuracy with different $T_{IOBB}$ on the ``old annotations'' test set on NIH.}
\label{tlb:acc_bb}
\begin{center}
\begin{tabular}{lcccc||c}
\hline
$T_{IOBB}$ & 0.1 & 0.25 & 0.5 & 0.75 & Avg\\
\hline\hline
Wang et al. \cite{wang2017chestx} & 0.15  & 0.05 & 0.00 & 0.00  & 0.04\\
Zhe et al. \cite{li2017thoracic} & \textbf{0.40}  & 0.25 & 0.11 & 0.07 & 0.18\\
RetinaNet: real  & 0.15  & 0.15 & 0.08 & 0.08 &  0.09 \\
RetinaNet: real + syn from CycleGAN + BboxLoss & 0.31 & 0.31 & 0.00 & 0.00 & 0.12
\\
RetinaNet: real + syn from DetectorGAN  & 0.31  & \textbf{0.31} & \textbf{0.23} & \textbf{0.23} & \textbf{0.22}\\

\hline
\end{tabular}
\end{center}
\end{table*}

\begin{table*}[htbp]
\centering
\caption{User study on the NIH Chest X-ray dataset.}
\label{tab:userstudy}
\begin{tabular}{lcccccccc}
\toprule
\multirow{2}{*}{Method} & \multicolumn{3}{c}{Prefer [\%] $\uparrow$} & \multicolumn{3}{c}{Mean Likert $\uparrow$} & \multicolumn{2}{c}{Std Likert} \\ 
\cmidrule{2-3}  \cmidrule{5-6}  \cmidrule{8-9}  
& Object & Whole & & Object & Whole & & Object & Whole \\
\midrule
CycleGAN + BboxLoss & 20 & 4 & & 1.31 & 1.18 & & 0.68 & 0.53 \\
DetectorGAN & \textbf{80} & \textbf{96} & & \textbf{2.69} & \textbf{3.88} & & 0.85 & 0.73 \\
\bottomrule
\end{tabular}
\end{table*}

\begin{figure*}[htbp]
\begin{center}
\includegraphics[width=1\linewidth]{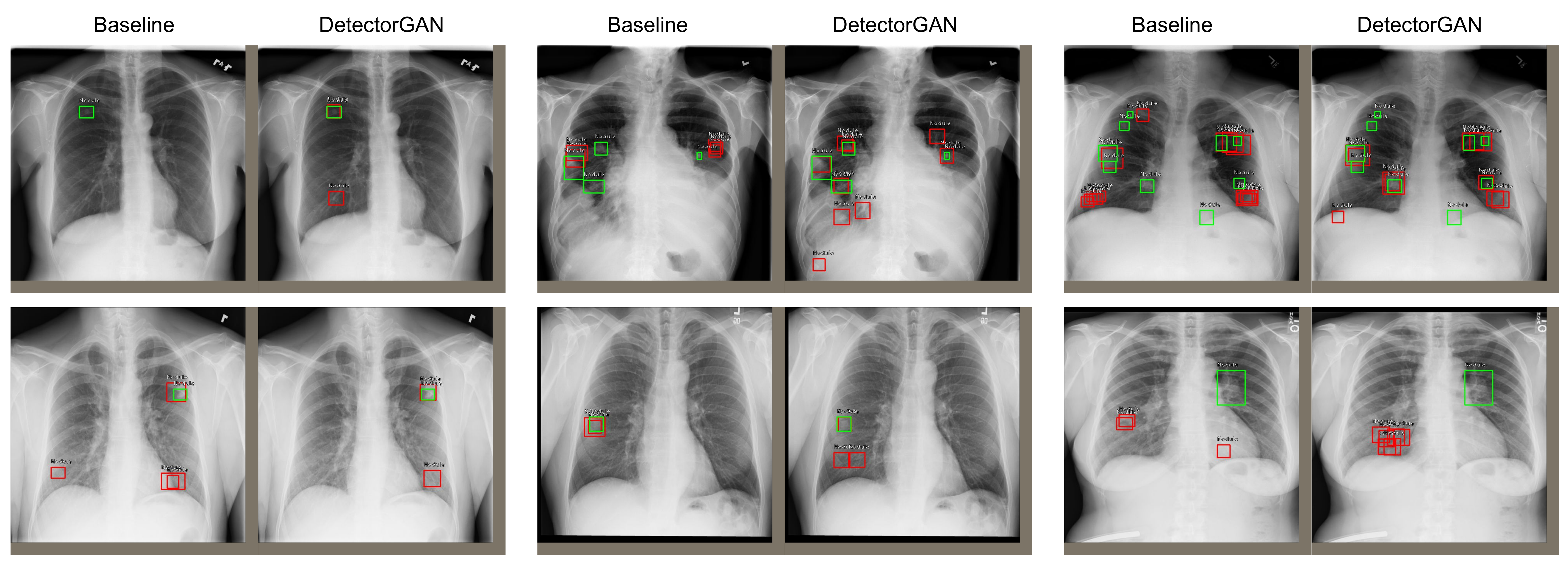}
\end{center}
   \caption{Comparison showing that adding synthetic images can help detect nodules in NIH Chest X-ray more accurately. Here, green boxes are ground truth and red are predictions. }
\label{fig:nodule_detected}
\end{figure*}

\section{Experiments}

In this section we demonstrate the effectiveness of DetectorGAN on two tasks: nodule detection task with the NIH Chest X-ray dataset and pedestrian detection with the Cityscapes dataset. 
We obtain significant improvements over baselines and achieve state of the art results on the nodule detection task.

\subsection{Disease Localization}

\subsubsection{Dataset}
We use the NIH Chest X-ray dataset \cite{wang2017chestx} and focus on the nodule detection task. 
The NIH Chest X-ray dataset contains 112,120 X-ray images -- 60,412 clean images and 51,708 disease images, 880 of which have bounding boxes. 
For the nodule class, there are 6,323 nodule images, 78 of which have bounding boxes.

\paragraph{Improved and Extended Annotations.}
The bounding box annotation for this dataset is however not satisfying due to the following issues: 
(1) In the original paper and previous work \cite{wang2017chestx,li2017thoracic}, there is no standard train/test/validation split. 
(2) The bounding box annotations are not complete; that is, for each image there is only at most one bounding box for each class annotated, while there are actually many nodules present in the image. 
(3) Even with a standard train/test/validation split, the test and validation sets are too small to obtain stable and meaningful results.

To address these problems, we make the following efforts to make the disease detection task more standard and easy to conduct research on:
(1) Generating a no-patient-overlap train/test/validation split with 0.7/0.2/0.1 portion of the data, yielding 57/13/9 images with 57/13/9 object instances.  
(2) Asking radiologists to re-annotate the current validation and test images using additional images from labeled images in the test/validation sets. 
These efforts result in 36 images and 80 images in validation and test sets accordingly, with 159 and 309 object instances. 
These splits and extended annotations will be published online to facilitate future research into this topic. 
We did not re-annotate or expand the training set as we want to demonstrate the effectiveness of the proposed method in learning small-data object detection tasks.

We refer to the 9/13 validation/test settings as ``old annotations'' and the 36/80 validation/test settings as the ``new annotations''. We obtain the detection AP on the ``new annotations'' and localization accuracy on the ``old annotations'' for fair comparison with previously published results.

\paragraph{Baselines and Previous Work.}
The baselines are: training with only real images, with additional synthetic images generated from CycleGAN and BboxLoss, and with additional synthetic images generated from ACGAN-like losses. 
The ACGAN-like losses refers to that in addition to discriminator losses, we also minimize the detection loss on synthetic images, similar to what ACGAN does for a classifier.
We compare these methods on the new high quality annotations.
In addition, we compare to two previously published best-performing works~\cite{wang2017chestx,li2017thoracic} using their evaluation split and their annotations (the ``old annotations'').

\paragraph{Evaluation Metrics.}
We use the standard object detection metric, average precision (AP), as the evaluation measure for the detection task. 
For comparisons to previous work, we also use their metric: localization accuracy, which is defined as the percentage of images that obtain correct predictions. 
An image is considered having correct predictions if the intersection over union ($IOU$) ratio between the predicted regions (can be non-rectangle) and the ground truth box is above threshold $T_{IOU}$. 
Another metric that is used by these works is to replace the $IOU$ with intersection over bounding boxes $IOBB$. 
However, we encourage researchers to use the proposed new annotations and evaluation metric in the future for standard comparisons.

\subsubsection{Quantitative Comparison}
\paragraph{New Annotation with Average Precision.}
In Table~\ref{tlb:ap}, we compare the results of using only real data, using synthetic data from the proposed method as well as from other baseline GAN models. 
We observe that DetectorGAN significantly improves the average precision. 
Compared to training on real data only, the AP nearly doubles from 0.124 to 0.236, and recall over triples from 0.184 to 0.649.
Compared to ACGAN-like losses and CycleGAN + BboxLoss, we obtain relatively 50\% and 20\% improvement.

We notice that ACGAN-like losses performs more poorly than using discriminator losses only, even though it has an additional loss to improve the detection performance on synthetic images. 
One explanation is that the generator and the detector learn only to detect synthetic objects, which is different from the goal of detecting real objects, leading to poor performance. 

To further demonstrate the benefits of using the unrolling step to bridge the gap between the generator and the detection performance on real images, we also experiment with a `DetectorGAN - unrolling' network without unrolling.  We observe a significant boost for adding the unrolling step, from 0.203 to 0.236 AP.

\paragraph{Old Annotation with Localization Accuracy.}
For comparison with previous work, we evaluate detection results using the localization accuracy metric with different $IOU$ and $IOBB$ thresholds. 
Results are shown in Table~\ref{tlb:acc_u} and Table~\ref{tlb:acc_bb}. 
We significantly outperform competing methods by relative 50\% and 22\%.

\subsubsection{Qualitative Analysis}
\paragraph{Generated Image Quality.}
We show DetectorGAN's generated images, along with CycleGAN-generated images in Fig.~\ref{fig:img_quality}. 
We observe that images are much better in terms of realism and blend-in.

\paragraph{Detected Nodules.}
We show that the detector helps to detect undetectable nodules in Fig.~\ref{fig:nodule_detected}.
We observe that every nodule captured by the baseline (trained on real images only) is also captured by the model trained using synthetic images. 
Meanwhile, adding synthetic images helps capture more nodules that baseline cannot capture.
Moreover, the box locations are generally more accurate.

\subsubsection{User Study}
We also conduct user study with a radiologist to evaluate the quality of the generated images. 
We ask the radiologists to rate the realism of the inserted nodule and the global image on a Likert scale (scale 1--5, with 5 indicating highest quality).
As shown in Table~\ref{tab:userstudy}, the images from DetectorGAN are better than those from CycleGAN + BboxLoss in 96\% of cases, with generated objects (nodules) better in 80\% of cases.
Moreover, the average Likert scores are significantly higher: 2.69 vs 1.31 for the objects, and 3.88 vs 1.18 for the whole image, demonstrating the benefits of our method.

\begin{table}[!h]
\caption{Pedestrian detection AP trained with real data, synthetic data generated by PS-GAN, pix2pix and DetectorGAN.}
\vspace{-3mm}
\begin{center}
\begin{tabular}{lcccc}
\hline
Data & Real & +DetectorGAN & +PS-GAN & +pix2pix \\
\hline
\hline
AP  &  0.593  & \textbf{0.613} & 0.602 & 0.574 \\
\hline

\end{tabular}
\end{center}
\label{tlb:ped_num}
\end{table}

\vspace{-2mm}

\subsection{Pedestrian Detection}
As a demonstration of the applicability of DetectorGAN to other datasets and problems, we apply it to pedestrian detection with a different base architecture.
We follow PS-GAN~\cite{ouyang2018pedestrian} to synthesize images with pedestrians inserted and improve pedestrian detection.
We demonstrate a quantitative and qualitative improvement in the generated images by adding the detector into the loop.

\paragraph{Dataset.}
We use the Cityscapes dataset, which contains 5,000 urban scene images with high-quality annotations.
We follow the instructions in the PS-GAN paper to filter images with small or occluded pedestrians obtain about 2,000 images with  about 9,000 labeled instances.

\paragraph{Baseline and Architecture.}
We use PS-GAN\cite{ouyang2018pedestrian} as the backbone architecture and add the detector into the model. The PS-GAN uses the standard pix2pix framework with local discriminators. 
This also shows that the DetectorGAN idea is versatile --- it can be integrated with different GAN models.
We fine-tune the model from the pretrained PS-GAN model.

\paragraph{Quantitative Results.}
Table~\ref{tlb:ped_num} shows that we improve the detection performance for pedestrian detection as well. 
We observe that DetectorGAN further improves the performance over PS-GAN.

All models here are trained using the same setting. 
The real-images-only baseline performance is slightly different from what is reported in the PS-GAN paper because we do not have access to the exact details of the detector setting used in the PS-GAN paper.

\paragraph{Qualitative Results.}
Qualitative results are shown in Fig.~\ref{fig:ped_img}.
We observe that DetectorGAN can generate qualitatively better images with less artifacts.

\section{Conclusion}

In this work we explored the object detection problem in the small data regime from a generative modeling perspective by learning to generate new images with associated bounding boxes.
We have shown that simply training an existing generative model does not yield satisfactory performance due to it optimizing for image realism instead of object detection accuracy.
To this end we developed a new model with a novel unrolling step that jointly optimizes a generative model and a detector such that the generated images improve the performance of the detector.
We show that this method significantly outperforms the state of the art on two challenging datasets.

{\small
\bibliographystyle{ieee_fullname}
\bibliography{bboxgan}
}

\end{document}